\documentclass[sigconf,nonacm]{acmart}

\copyrightyear{2027}
\acmYear{2027}
\setcopyright{acmlicensed}
\acmConference[WSDM '27]{The 20th ACM International Conference on Web Search and Data Mining}{February 15--19, 2027}{Hong Kong}
\acmBooktitle{WSDM '27, February 15--19, 2027, Hong Kong}
\acmPrice{}
\acmDOI{}
\acmISBN{}

\usepackage{booktabs}
\usepackage{graphicx}

\newcommand{\Nagents}{2{,}617}      
\newcommand{\Nanalysis}{2{,}499}    
\newcommand{\Ngen}{2{,}894}         
\newcommand{\rhohat}{0.92}
\newcommand{\dbic}{51.8}
\newcommand{\gapmain}{0.71}
\newcommand{\gappartial}{0.69}

\begin{document}

\title{Delegation Asymmetry in Agentic Recommender Systems: Measuring Two-Sided Receptivity in Online Dating}

\author{Daria Leshchikova}
\affiliation{\institution{Fleamily, Inc.}\city{Delaware}\country{USA}}
\author{Valentina V. Kuskova}
\affiliation{\institution{Lucy Family Institute for Data \& Society \\ University of Notre Dame}\city{Notre Dame, IN}\country{USA}}
\author{Dmitry Zaytsev}
\affiliation{\institution{Lucy Family Institute for Data \& Society \\ University of Notre Dame}\city{Notre Dame, IN}\country{USA}}
\author{Valerii Klimov}
\affiliation{\institution{Fleamily, Inc.}\city{Delaware}\country{USA}}

\begin{abstract}
Autonomous LLM agents that converse on a user's behalf are an emerging design pattern in matching platforms, yet their viability depends on a condition rarely examined: users must accept not only \emph{delegating} conversation to an agent, but also \emph{receiving} agent-mediated communication from others. We study this condition using two large-scale surveys of active users of a major dating platform (N~=~\Ngen{} on generative profile features; N~=~\Nagents{} on autonomous conversational agents, fielded in two languages). We develop a latent-variable measurement model of agent receptivity based on graded response models with latent regression, and show via model comparison that willingness to send and willingness to receive agent communication are distinct constructs. They are highly correlated ($\rho = \rhohat$) but separable ($\Delta\mathrm{BIC} = 52$), with partial measurement invariance across languages. The model quantifies a systematic \emph{delegation asymmetry}: deploying one's own agent requires far lower receptivity (threshold $\theta = -0.38$) than engaging a counterpart's agent ($\theta = +0.32$; full engagement $+1.39$), and mean deployment propensity exceeds engagement propensity roughly threefold. Under a random-pairing counterfactual derived from stated receptivity, only 4--13\% of directed dyads combine agent deployment with receiver engagement, with a pronounced gender-directional imbalance. Design counterfactuals quantify the levers: a reciprocity requirement halves interaction volume by excluding two-thirds of would-be deployers, while routing agent contacts on receive receptivity triples per-contact engagement: a lift that survives out-of-sample validation in which the routing score is estimated without the target item (AUC 0.88, 3.1$\times$ quartile lift under respondent-level cross-validation). We discuss implications for agentic recommender design, including disclosure, opt-in mechanics, and receptivity-aware matchmaking.
\end{abstract}



\maketitle

\section{Introduction}
Somewhere tonight, a dating-app user will get a warm, well-crafted opening message from a match: composed, sent, and if things go well, followed up on by a language model while its principal sleeps. The scenario is no longer speculative. Recent survey evidence suggests that more than a quarter of U.S. singles have already used AI somewhere in their dating lives, and a majority of dating-app users believe they have, at some point, exchanged messages with an algorithm rather than a person \cite{ante2026cyrano}. A cottage industry of ``AI wingman'' tools writes openers and replies on demand; users who receive such messages report feeling deceived when they find out \cite{ante2026cyrano}, and merely \emph{suspecting} AI authorship is enough to depress trust \cite{jakesch2019ai,kadoma2025generative}. What began as Cyrano de Bergerac with a keyboard - one party quietly borrowing better words - is being industrialized.

Platforms are now productizing the next step. The first wave of generative features was \emph{assistance}: AI-written profile summaries, suggested openers, tone polish, with the user remaining the author of record. The emerging wave is \emph{delegation}: autonomous LLM agents that hold conversations on the user's behalf are initiating contact, exchanging messages, screening counterparts, even negotiating common ground with the other side's agent before the humans ever speak. The research community is actively building this machinery \cite{peng2025survey,zhang2024agentcf,wang2024recmind,ye2025cognipair}, and online dating is its natural frontier: communication is not a feature of the product but the product itself, interaction volume is high, and the stakes are as personal as software gets.

There is, however, a condition for this market that the building has run ahead of. Agent-mediated matching is a \emph{two-sided} proposition: every delegated message needs a willing receiver on the other end. A market in which many users deploy agents but few will engage with one is not a matching market at all; it is automated outreach landing on closed doors, at scale, in the most trust-sensitive consumer domain there is. Yet the two sides of this condition have never been measured together. A large literature measures acceptance of AI tools by their \emph{users} \cite{davis1989perceived,venkatesh2003user}, and a growing one documents receiver-side penalties when AI involvement is suspected \cite{jakesch2019ai,hancock2020ai}; what is missing is a joint, within-person measurement of willingness to \emph{send} and willingness to \emph{receive} agent communication on a common scale - the quantity that determines whether both sides of a delegation market exist, and the quantity this paper provides.

We measure both sides. Working with two large surveys fielded to active users of a major dating platform (N~=~\Ngen{} on generative profile features; N~=~\Nagents{} on autonomous conversational agents, in two languages), we build a latent-variable measurement model of \emph{agent receptivity}: a graded response model with latent regression in which seven attitudinal items load on a \emph{send} dimension (willingness to delegate one's own communication) and a \emph{receive} dimension (willingness to engage with agent-mediated communication from others).

Four findings emerge. First, model comparison shows send and receive receptivity are \emph{distinct constructs}: correlated at $\rho = \rhohat$ but decisively separable ($\Delta\mathrm{BIC} = \dbic$), with partial measurement invariance across the two languages. Second, the model quantifies a \emph{delegation asymmetry}: on the common scale, deploying one's own agent requires receptivity $\theta = -0.38$ while engaging a counterpart's agent requires $\theta = +0.32$, a displacement of \gapmain{}~SD that survives every recoding we test. With that, population deployment propensity exceeds engagement propensity roughly threefold. Third, receptivity is \emph{associated with unmet need}: it concentrates among users whose matches fail to convert to conversations and who report frustration about stalled matches. It is lower among women and long-tenured users, and the gender gap is a trait difference, not an artifact of differential item functioning. Fourth, under a random-pairing stated-preference counterfactual, only 4--13\% of directed dyads combine deployment with engagement, with a pronounced gender-directional imbalance; counterfactuals quantify the design levers available to a platform.

This paper makes four contributions. \textbf{The first is a construct and a measurement design.} We introduce \emph{send receptivity} and \emph{receive receptivity} as distinct constructs and measure them jointly, within person, on a common latent scale. Prior work compares AI users and AI receivers across separate studies and samples, confounding population differences with role differences. The within-person design makes the send--receive asymmetry observable without confounding it with between-sample population differences. 

\textbf{The second is receptivity audit, a general methodology.} We package the pipeline, from short ordinal instrument $\to$ latent measurement model $\to$ model-implied per-user deploy/engage endorsement propensities $\to$ out-of-sample validation $\to$ two-sided market counterfactuals as a reusable, pre-deployment procedure (Section ~\ref{sec:audit}). It tells a platform whether both sides of an agent-mediated feature's market exist \emph{before the agents are built}, and hands the matching layer a pre-deployment routing signal, validated out of sample. Nothing in the procedure is specific to dating. Its methodological content is the bridge: item-level psychometric parameters used directly as market primitives for agentic-system design.
 
\textbf{The third is empirical findings at platform scale.} On a major dating platform we establish that send and receive receptivity are distinct ($\rho = \rhohat$, $\Delta\mathrm{BIC} = \dbic$), quantify the delegation asymmetry (\gapmain{}~SD, CI $[0.65,0.77]$, robust to recodings and partial measurement invariance), show receptivity concentrates among users reporting unmet matching need and is gendered, and identify a stable asymmetric-delegator segment ($\approx$25\%). The asymmetry is visible without any model: 40.7\% of respondents individually rate sending above receiving; 2.1\% the reverse. 

\textbf{The fourth is market consequences and quantified design levers.} Under a random-pairing counterfactual on the fitted stated-preference model, we price the design space: baseline viability of 4--13\% of directed dyads, a gender-directional imbalance, the cost of a reciprocity norm ($-$65\% of deployers), and a receptivity-aware routing frontier whose 3$\times$ per-contact engagement gain survives leave-target-item-out cross-validation, with receive-side consent emerging as a first-class design primitive.

\section{Related Work}

\textbf{Online dating as a matching market.}
Online dating is now the dominant channel through which couples meet in several countries \cite{rosenfeld2019disintermediating}, and its market structure is well documented: users sort on observable attributes with strong revealed preferences \cite{hitsch2010matching}. As users direct most of their attention upward toward more desirable counterparts \cite{bruch2018aspirational}, most first messages go unanswered: reply behavior is scarce, skewed, and predictable from user and dyad features \cite{xia2014predicting}. The domain's relationship psychology is surveyed by Finkel et al.~\cite{finkel2012online}. This attention-scarcity structure is the backdrop for our results: automated delegation promises to multiply first-message volume precisely where reply willingness is already the binding resource, and our receive-side measurements quantify how much colder the reception gets when the message is agent-authored.

\textbf{Reciprocal recommendation.}
Algorithmically, dating recommenders differ from item recommenders in requiring \emph{mutual} interest. The reciprocal-recommendation line runs from RECON's harmonic combination of directional preferences \cite{pizzato2010recon,pizzato2013recommending} through behavioral models on large platforms \cite{xia2015reciprocal}, latent-factor formulations \cite{neve2019latent}, and the survey of Palomares et al.~\cite{palomares2021reciprocal}. Recent work treats matching platforms explicitly as two-sided markets with exposure trade-offs: fairness-aware reciprocal recommendation \cite{tomita2024fair}, welfare-based balancing of match rates and fairness, and off-policy evaluation for matching interventions \cite{hayashi2025off}; agent-based models reproduce engagement cycles in swiping apps from simple behavioral rules \cite{cela2025emotional}. Users, for their part, hold rich folk theories about dating algorithms and counter-strategize against perceived platform interests \cite{alizadeh2024matchmaker}. This literature optimizes \emph{who is shown to whom}, taking the communication medium as given; agent mediation makes the medium itself a market variable, adding a second reciprocity layer:  both sides must accept not only each other but the \emph{mode} of contact. Our send/receive distinction formalizes this second layer, and our routing counterfactual (\S\ref{sec:market}) is a reciprocal-recommendation intervention on it.

\textbf{AI-mediated communication.}
Hancock et al.~\cite{hancock2020ai} define AI-mediated communication and its research agenda. Receiver-side penalties are well documented: profiles suspected of AI authorship are trusted less \cite{jakesch2019ai}, suspicion of LLM use falls unevenly across writers \cite{kadoma2025generative}, and smart replies measurably alter language and relationships \cite{hohenstein2023artificial}. Closest to our setting, Ante~\cite{ante2026cyrano} interviews both LLM-assisted senders and receivers in online dating, finding an ``authenticity paradox'' among senders and ``digital betrayal'' among receivers. These qualitative asymmetries motivate our contribution: measuring send- and receive-willingness jointly, within person, on one latent scale, at population size.

\textbf{LLM agents and agentic recommenders.}
LLM agents that plan, remember, and act \cite{park2023generative} are moving into recommendation: as autonomous user simulacra for collaborative filtering \cite{zhang2024agentcf}, as planning agents over recommender tools \cite{wang2024recmind}, as user-side shields that renegotiate the user--platform relationship \cite{xu2025iagent}, and as multi-agent matching prototypes for dating specifically \cite{ye2025cognipair}; Peng et al.~\cite{peng2025survey} survey the space. This literature builds the supply side of agent-mediated interaction. Our results measure the demand side and show the receiving half of that demand is the binding constraint.

\textbf{Measuring attitudes toward AI.}
Stated-preference measurement of technology adoption descends from TAM and UTAUT \cite{davis1989perceived,venkatesh2003user}. For AI specifically, experimental work established that people both discount algorithmic judgment after seeing it err \cite{dietvorst2015algorithm} and, in other conditions, prefer it to human judgment \cite{logg2019algorithm}. This is a polarity that our rejector/enthusiast class structure recovers at population scale. A family of dedicated self-report instruments has followed: the two-factor GAAIS \cite{schepman2020initial,schepman2023general}, the brief AIAS-4 \cite{grassini2023development}, and the unidimensional ATTARI-12 \cite{stein2024attitudes}. These scales share three properties that limit them for our question: they measure attitudes toward AI \emph{in general} rather than toward a concrete interpersonal deployment; they are sum-scored composites, so response-category thresholds are not modeled quantities; and they measure the respondent as a \emph{user} of AI, with no construct for being on the receiving end of someone else's AI. Our instrument and model address all three: concrete agent scenarios, threshold-level modeling, and joint within-person measurement of send and receive dispositions.

\textbf{Psychometric machinery.}
We use item response theory \cite{embretson2000item}: the graded response model for ordinal items \cite{samejima1969estimation}, estimated by marginal maximum likelihood \cite{bock1981marginal}, extended with latent regression on covariates and formal invariance testing in the factorial-invariance tradition \cite{meredith1993measurement,millsap2011statistical}. The same model family aggregates ordinal expert ratings into latent indices in cross-national measurement programs \cite{pemstein2018vdem}, where, in the same way as here, the scientific claims live in threshold locations and cross-group comparability rather than in raw scores. The payoff for us is direct: the delegation asymmetry is a statement about where two thresholds sit on one latent scale, which requires thresholds to be first-class parameters rather than artifacts of scale construction, and the partial-invariance analysis (\S\ref{sec:results}) is the standard machinery for licensing our cross-language pooling.

\section{The Receptivity Audit}
\label{sec:audit}
Agent-mediated communication features pose a question that arrives \emph{before} any system exists to log behavior from: do both sides of the market for this feature exist? Answering it post-launch is expensive: the failure mode is deployed agents generating unwanted contact at scale in a trust-sensitive domain. Answering it with off-the-shelf attitude scales is not possible, because existing instruments measure attitudes toward AI in general, score respondents as \emph{users} only, and collapse responses into sums that cannot locate thresholds (\S2). We therefore formalize the procedure this paper instantiates as a general, four-stage \emph{receptivity audit}:

\begin{enumerate}
\item \textbf{Instrument.} A short battery of ordinal items presenting the concrete feature to its intended population, with items deliberately spanning both roles: scenarios in which the respondent \emph{deploys} the capability, and scenarios in which the respondent is \emph{on the receiving end} of someone else's deployment. Within-person coverage of both roles is the non-negotiable design element; it is what separates role effects from population effects.
\item \textbf{Measurement model.} A confirmatory multidimensional graded response model with latent regression: items load on send and receive factors, latent traits regress on user covariates, and dimensionality itself is a model-comparison question rather than an assumption. The fitted model yields three things no sum score provides: threshold locations on a common scale (the asymmetry is a statement about thresholds), formal invariance tests licensing pooling across languages or segments, and per-user factor scores with principled uncertainty.
\item \textbf{Propensities, validated out of sample.} Model-implied \emph{endorsement} propensities computed from the fitted item response functions at each user's factor scores, such as the probability of endorsing deployment, and the probability of endorsing engagement when contacted, under transparent strict/soft operationalizations of intermediate response categories, treated as scenario bounds rather than calibrated behavior. Before the propensities feed any policy, the score behind them is validated predictively: the target item is held out, the score is re-estimated from the remaining items and covariates under respondent-level cross-validation, and out-of-sample discrimination and lift are reported. This is what licenses using the score as a routing signal.
\item \textbf{Market counterfactuals.} A dyadic simulation over the empirical population that converts propensities into market quantities such as interaction volume, engagement per contact, directional imbalances, and prices design levers - eligibility rules (who may deploy), routing rules (who receives), and disclosure or consent mechanics. Because the levers are expressed in the model's own parameters, each counterfactual is a closed-form computation, not a new experiment.
\end{enumerate}

Two remarks on scope. First, the audit is deliberately parsimonious: every component we use is established machinery \cite{samejima1969estimation,bock1981marginal,meredith1993measurement}; the contribution we claim is the construct and the bridge to market primitives, not a new estimator. Second, nothing above mentions dating. The audit applies wherever one side delegates communication to an agent and the other side must accept the medium: recruiting outreach, sales prospecting, agent-mediated customer contact, professional networking. Sections~\ref{sec:data}--\ref{sec:market} instantiate the four stages on one such market; \S\ref{sec:beyond} returns to the general case.

\section{Data and Instruments}
\label{sec:data}
Two survey instruments were fielded to active users of a large dating platform, \href{https://fledge.love}{Fledge.Love} as voluntary, self-administered online questionnaires. Respondents were recruited through an in-app prompt shown to active users. The agents instrument (B) was fielded first, from November 12 to December 15, 2025, as parallel Russian- and English-language forms collected concurrently. The generative-features instrument (A) followed from March 22 to April 1, 2026, in Russian. The two samples are not linked at the respondent level.

\textbf{Instrument A: generative features (N = \Ngen).}
Respondents were shown three generative-AI feature concepts in sequence, each with a short description: (1) an \emph{AI profile summary}: a generated synopsis of the user's personality-test results displayed on their profile; (2) \emph{AI conversation tips}: generated suggestions for messaging a specific match; and (3) an \emph{AI couple description}: a generated compatibility narrative for a user--match pair. Each concept was rated on a three-level interest scale (\emph{not interesting} / \emph{interesting} / \emph{very interesting}), with an open ``other'' field that a minority of respondents used for free-text commentary instead of a rating (n = 473 comments across items; these respondents are excluded from the corresponding item's denominator and their comments reviewed for context but not analyzed further). Respondents then chose the best of the three concepts (with a ``none'' option). They further answered whether such features would increase their interest in browsing profiles (four levels from \emph{definitely yes} to \emph{engagement would not increase}), alongside gender, age band, platform-usage frequency, and everyday AI usage. The sample (N = \Ngen) is 59.7\% male; 25.3\% aged 18--24, 40.1\% aged 25--34, 25.5\% aged 35--44, and 9.1\% 45+. It skews toward engaged users:  80.1\% report opening the app several times a week, and toward AI-familiar ones: 47.2\% report using AI services regularly and a further 29.4\% occasionally. A contact field completed by 1{,}037 respondents volunteering for follow-up interviews is excluded from all analysis and from the deposit.

\textbf{Instrument B: autonomous agents (N = \Nagents; RU and EN).}
The second instrument presented a concrete agent feature through a sequence of annotated interface mockups: a conversational agent that messages the user's matches on their behalf, with user-configurable activity level (initiates vs.\ replies only) and tone. Items were interleaved with the mockups so that each scenario was rated immediately after being shown. The seven attitudinal items, their roles, and their response options are given in Table~\ref{tab:instrument}; English wording is shown, with the Russian form a professional parallel translation. The design deliberately spans both roles of the delegation relation: Y1--Y3 and Y7 place the respondent as the agent's \emph{principal} (seeing the concept, configuring the agent, delegating one's own conversations, valuing the product), while Y4--Y6 place the respondent as the \emph{counterpart} (receiving messages from someone else's agent, observing agent-to-agent pre-conversation about oneself, sharing a group chat with agents). This within-person role coverage is stage one of the audit (\S\ref{sec:audit}).

The instrument also captured six covariates on ordinal scales: gender; age band (18--24 / 25--34 / 35--44 / 45+); platform tenure (first month / several months / over a year); perceived match volume (very few / not many / enough / too many); match-to-conversation conversion (very few / less than half / more than half / almost all); and negative affect about stalled matches (none / neutral / yes). Two survey questions on conversation-decay attributions and coping strategy are not used in the measurement model.

\emph{Coding.} Responses are coded so that higher categories indicate greater receptivity. Two items required ordering judgments: Y1 offers two negative options (\emph{not interesting}, \emph{didn't seem trustworthy}) that we collapse into a single lowest category, and Y4 similarly offers two rejections (\emph{would react negatively}, \emph{weird  - only want a real person}); Y7's 1--10 rating is binned into five ordered levels. Section~\ref{sec:results} reports sensitivity analyses in which the collapsed categories are instead treated as distinct ordered levels; the headline quantities are stable (asymmetry gap 0.70--0.72 across codings). The full codebook, including both language forms, ships with the released pipeline. Complete cases across items and covariates: N = \Nanalysis{} (95.5\% of responses), of which 212 are from the EN form.

\textbf{The released dataset.} The deposit contains one record per response, in two files mirroring the instruments. The Instrument~B file (2{,}617 records: 2{,}385 RU, 232 EN) carries: a language flag; the seven ordinal items of Table~\ref{tab:instrument} both as verbatim categorical responses and under the analysis coding; the six ordinal covariates (gender, age band, tenure, match volume, match-to-conversation conversion, negative affect about stalled matches); and two auxiliary categorical fields not used in the measurement model (attributed reason for conversation decay; coping strategy for stalled matches). The Instrument~A file (2{,}894 records) carries: the three concept ratings on the three-level scale (with free-text ``other'' responses replaced by a nonresponse flag); the best-of-three choice; the browsing-interest item; and gender, age band, usage frequency, and AI-usage frequency. An auxiliary file provides, for each of the 2{,}499 complete cases, the model-derived quantities used in Sections~\ref{sec:results}--\ref{sec:market}: EAP factor scores $\hat\theta^{\mathrm{send}}, \hat\theta^{\mathrm{recv}}$ and the four deploy/engage propensities under strict and soft operationalizations. Removed relative to the raw collection: contact information, all free-text (473 comments in Instrument~A, reviewed for context only and not released), and exact submission timestamps, which are coarsened to the ISO week; demographic cross-classifications were audited so that no released cell isolates fewer than $k = 5$ respondents (binding only in the small EN subsample). A codebook documents both language forms verbatim, every coding map, and the provenance of each derived field. The released files are the analysis dataset of record: the $k$-anonymity recoding (which alters the age code of three EN respondents) was applied \emph{before} the final model fits, so every quantity in this paper reproduces exactly from the deposit.

\textbf{Human-subjects review.} Both surveys were designed, fielded, and collected by the platform for product research; the authors received the response data as an existing dataset and conducted secondary analysis only. The analysis protocol was reviewed by the authors' institutional review board and determined not to constitute human subjects research, as secondary analysis of de-identified data (protocol 26-08-10287).

\textbf{Availability.} Data, codebook, and the full analysis pipeline (harmonization, model fitting, robustness battery, simulation, and table generation) are archived at DOI  \href{ https://zenodo.org/records/21971273}{\texttt{10.5281/zenodo.21971273}} . Every number, table, and figure in this paper is reproducible from the deposit via the released scripts.

\begin{table*}[t]
\caption{Instrument B: agent-receptivity items. English form shown; higher codes indicate greater receptivity. P = respondent as principal (send dimension), C = respondent as counterpart (receive dimension).}
\label{tab:instrument}
\small
\begin{tabular}{clp{6.2cm}p{7.2cm}}
\toprule
Item & Role & Stem (abridged) & Response options (code) \\
\midrule
Y1 & P & First emotional reaction to the agent concept & very interesting (2); neutral (1); didn't seem trustworthy (0); not interesting (0) \\
Y2 & P & Attitude to configuring the agent's activity and tone & useful -- saves me time (2); too complicated -- can't be bothered (1); risky -- might write something inappropriate (0) \\
Y3 & P & ``How would you feel if the agent talked with a potential partner while you are away?'' & I want to try it (2); I might try it (1); I don't want an agent responding for me (0) \\
Y4 & C & ``How would you feel if someone's AI agent responded to you?'' & would engage in conversation with the agent (2); might engage (1); weird -- only want a real person (0); would react negatively (0) \\
Y5 & C & Your agent chats with another agent to find common interests before you join & super idea (3); funny -- would watch their conversation (2); doubtful -- why are we needed then? (1); strange and frightening (0) \\
Y6 & C & Humans and agents converse together in a group chat & super idea (2); funny -- would participate (1); doubtful / strange (0) \\
Y7 & P & Overall value of the feature for staying active and not losing matches, 1--10 & binned: 1--2 (0); 3--4 (1); 5--6 (2); 7--8 (3); 9--10 (4) \\
\bottomrule
\end{tabular}
\end{table*}

\begin{table}[t]
\caption{Sample descriptives, agent-survey complete cases (N = 2{,}499).}
\label{tab:descriptives}
\small
\begin{tabular}{lr}
\toprule
Female & 35.4\% \\
Age 18--24 & 21.1\% \\
Age 25--34 & 40.8\% \\
Age 35--44 & 29.3\% \\
Age 45+ & 8.8\% \\
Tenure: $<$1 month & 24.6\% \\
Tenure: Several months & 44.5\% \\
Tenure: $>$1 year & 30.9\% \\
EN-language instrument & 8.5\% \\
\bottomrule
\end{tabular}

\end{table}

\section{A Measurement Model of Agent Receptivity}
\label{sec:model}
We model the seven ordinal items with a graded response model \cite{samejima1969estimation} with latent regression. Person $i$'s response to item $j$ with $K_j$ ordered categories follows
\begin{equation}
\Pr(Y_{ij} \ge k \mid \boldsymbol\theta_i) = \sigma\!\left(a_j(\theta_{i,d(j)} - b_{jk})\right),
\end{equation}
with discrimination $a_j > 0$ and ordered thresholds $b_{j1} < \dots < b_{j,K_j-1}$. In the two-dimensional specification, $d(j)$ assigns items to a \emph{send} factor (Y1, Y2, Y3, Y7) or a \emph{receive} factor (Y4, Y5, Y6); the latent traits follow
\begin{equation}
\boldsymbol\theta_i = B^{\top} \mathbf{x}_i + \boldsymbol\varepsilon_i, \qquad
\boldsymbol\varepsilon_i \sim \mathcal{N}\!\left(\mathbf{0}, \begin{bmatrix} 1 & \rho \\ \rho & 1 \end{bmatrix}\right),
\end{equation}
where $\mathbf{x}_i$ collects centered covariates (gender, age, tenure, match volume, conversion, negative affect, language).

\textbf{Why measurement models outperform sum scores.} We treat receptivity as a latent variable in the structural tradition \cite{bollen1989structural,bollen2002latent,borsboom2003theoretical}: the items are fallible ordinal indicators of unobserved dispositions, and the quantities of scientific interest are parameters of the measurement model, not arithmetic on raw responses. Sum scoring is not a model-free alternative: it is itself a latent variable model with strict implicit constraints, notably equal item weights \cite{mcneish2020thinking}, and those constraints are untenable here on three grounds. The items have heterogeneous category counts (three to five levels), so equal weighting is arbitrary by construction. The scientific claim at stake, that engaging a counterpart's agent requires more receptivity than deploying one's own, is a claim about \emph{where response-category thresholds sit on a common latent scale}, and thresholds simply do not exist as quantities in a composite score. And pooling a two-language sample requires the formal invariance machinery of the factorial-invariance tradition \cite{meredith1993measurement,millsap2011statistical}, including the partial-invariance fallback we end up needing, which is native to latent variable models and unavailable to composites. The graded response model, a generalized latent variable model for ordinal indicators \cite{skrondal2004generalized}, delivers all three properties at the cost of distributional assumptions that we subject to model comparison rather than assume.

\textbf{Estimation and inference.} We estimate by marginal maximum likelihood \cite{bock1981marginal} with $13^2$-node Gauss--Hermite quadrature over the correlated latent space, optimizing with gradient-based quasi-Newton methods under automatic differentiation. Identification fixes both trait variances to one and centers all covariates, so thresholds absorb location and the latent regression is a pure slope structure. Dimensionality (1D vs.\ 2D) and the latent correlation are selected by BIC and likelihood-ratio test; per-user factor scores are expected a posteriori (EAP) values under the fitted model; uncertainty for all reported quantities comes from a 200-replicate nonparametric bootstrap over respondents (replicates refit at reduced quadrature for tractability; percentile intervals are recentered on the full-precision estimates); and differential item functioning is tested per item by likelihood ratio with Bonferroni correction, with a partial-invariance refit verifying that substantive conclusions survive freeing flagged items. Code for the full pipeline is released for reuse in the deposit described in \S\ref{sec:data}.

\section{Results: The Structure of Receptivity}
\label{sec:results}

\subsection{Send and receive are distinct constructs}
\begin{table}[t]
\caption{Model comparison. The two-dimensional model is preferred.}
\label{tab:comparison}
\small
\begin{tabular}{lrrr}
\toprule
Model & $-\log L$ & Params & BIC \\
\midrule
1D & 13,707.4 & 31 & 27,657.2 \\
2D & 13,650.1 & 39 & \textbf{27,605.4} \\
\bottomrule
\end{tabular}

\end{table}
The two-dimensional model dominates ($\Delta\mathrm{BIC} = \dbic$; LRT $= 114.5$, $8$~df, $p < 10^{-15}$), with latent correlation $\rho = \rhohat$: willingness to send and willingness to receive agent communication are strongly related but not interchangeable. Table~\ref{tab:items} reports the item parameters. All seven items discriminate strongly ($a$ between 1.7 and 4.2); the single most informative item on the send dimension is willingness to deploy one's own agent (Y3, $a = 4.2$), and the receive-side items discriminate comparably well. The weakest item is the overall 1--10 product rating (Y7, $a = 1.7$), whose thresholds sit far to the right of every other send-side item: even users well above average in receptivity stop short of rating the product concept highly. Curiosity about agents and perceived product value are related but partially decoupled, which is a caution against reading feature-level enthusiasm off a single product rating, in either direction.
\begin{table}[t]
\caption{Item parameters of the 2D graded response model: discrimination $a$ and ordered thresholds $b_k$ (the latent-trait location at which the probability of responding in category $k$ or higher reaches 50\%). Item content and response categories are given in Table~\ref{tab:instrument}; dashes mark items with fewer categories; S = send, R = receive dimension.}
\label{tab:items}
\small
\begin{tabular}{lcrrrrr}
\toprule
Item & Dim. & $a$ & $b_1$ & $b_2$ & $b_3$ & $b_4$ \\
\midrule
Y1 & S & 2.97 & $-0.23$ & $0.36$ & --- & --- \\
Y2 & S & 3.44 & $-0.33$ & $-0.12$ & --- & --- \\
Y3 & S & 4.20 & $-0.38$ & $0.38$ & --- & --- \\
Y4 & R & 2.82 & $0.32$ & $1.39$ & --- & --- \\
Y5 & R & 3.28 & $-1.64$ & $-0.36$ & $0.99$ & --- \\
Y6 & R & 3.36 & $-0.09$ & $1.17$ & --- & --- \\
Y7 & S & 1.72 & $0.48$ & $0.94$ & $1.53$ & $2.09$ \\
\bottomrule
\end{tabular}

\end{table}

\subsection{The delegation asymmetry}
\begin{figure}[t]
\includegraphics[width=\linewidth]{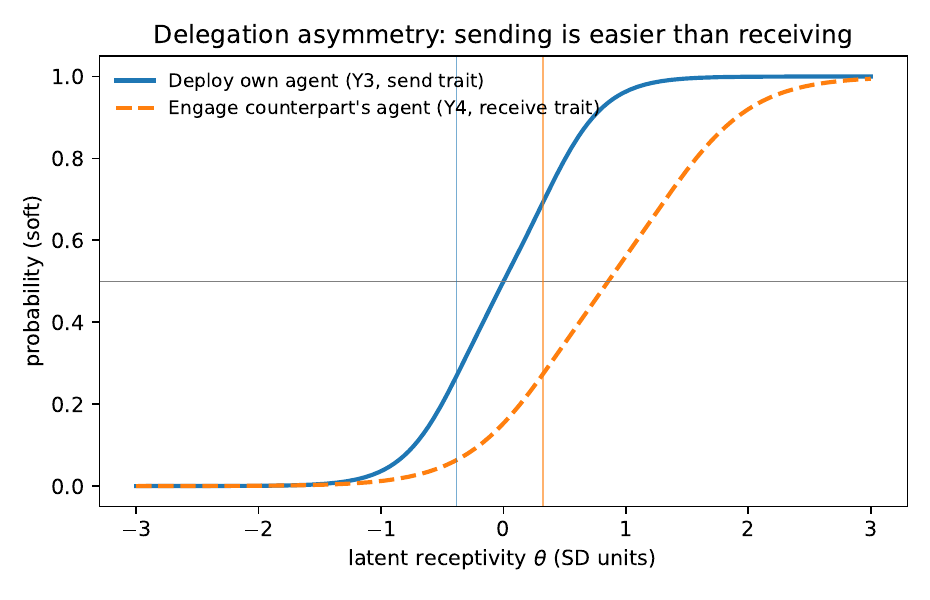}
\caption{Item response curves for deploying one's own agent (Y3) and engaging a counterpart's agent (Y4). The engagement curve is displaced $\approx$0.7~SD to the right.}
\label{fig:asymmetry}
\end{figure}
On the anchored latent scale, endorsing deployment of one's own agent requires $\theta = -0.38$, while endorsing engagement with a counterpart's agent requires $\theta = +0.32$ (full engagement $+1.39$), a displacement of \gapmain{}~SD (95\% nonparametric bootstrap CI $[0.65, 0.77]$, 200 replicates; Figure~\ref{fig:asymmetry}). The displacement is stable across the alternative category orderings we test ($0.70$--$0.72$), and the latent correlation is estimated precisely ($\rho = 0.92$, CI $[0.90, 0.94]$). Mean model-implied deployment endorsement is 0.38 (strict) / 0.50 (soft) against engagement endorsement of 0.12 / 0.26.

\emph{The asymmetry is not a modeling artifact.} Y3 and Y4 differ in wording and response options as well as role, so the threshold displacement could in principle absorb formulation differences. Three checks show the asymmetry is not an artifact of dimensional specification or coding, and is visible directly within respondents. 
\begin{itemize}
\item Model-free paired evidence: on the collapsed three-category codings, 40.7\% of respondents individually rate deploying their own agent strictly above engaging a counterpart's, against 2.1\% in the opposite direction: a 19:1 imbalance (Wilcoxon signed-rank $p < 10^{-189}$; sign test $p < 10^{-232}$). The modal cell of the paired table is joint rejection (36.1\%), and full send-endorsement coexists with full receive-endorsement in only 11.0\% of respondents. 
\item Dimensionality: refitting with all seven items on a \emph{single} latent dimension, so both thresholds live on literally the same scale, yields a displacement of $0.68$~SD, within a few hundredths of the 2D estimate. 
\item Coding: the displacement is stable ($0.70$--$0.72$) across the alternative category orderings of \S\ref{sec:data}. 
\end{itemize}
These checks cannot, however, separate a pure role effect from residual scenario-framing differences between the two stems (``while you are away'' vs.\ ``responded to you''); a paired vignette experiment randomizing role against fixed wording is the clean follow-up.

\subsection{Receptivity is associated with unmet need}
\begin{table}[t]
\caption{Latent regression of send and receive receptivity on centered covariates (trait SD = 1). Brackets: 95\% nonparametric bootstrap CIs (200 replicates over respondents), recentered on the full-precision estimates.}
\label{tab:regression}
\small
\begin{tabular}{lrr}
\toprule
Covariate & $\beta_{\mathrm{send}}$ & $\beta_{\mathrm{recv}}$ \\
\midrule
Female & $-0.38$ [-0.57, -0.23] & $-0.30$ [-0.47, -0.18] \\
Age band & $+0.07$ [0.00, 0.15] & $-0.01$ [-0.07, 0.06] \\
Platform tenure & $-0.14$ [-0.24, 0.00] & $-0.11$ [-0.19, 0.04] \\
Match volume & $+0.09$ [-0.04, 0.17] & $+0.08$ [-0.02, 0.14] \\
Match$\to$conversation rate & $-0.17$ [-0.27, -0.11] & $-0.16$ [-0.26, -0.09] \\
Negative affect (stalled matches) & $+0.14$ [0.07, 0.26] & $+0.14$ [0.06, 0.23] \\
Language: EN & $+0.01$ [-0.14, 0.16] & $+0.07$ [-0.09, 0.22] \\
\bottomrule
\end{tabular}

\end{table}
Table ~\ref{tab:regression} reports the latent regression. Women are substantially less receptive on both dimensions (send $\beta = -0.38$, CI $[-0.57, -0.23]$; receive $\beta = -0.30$, CI $[-0.47, -0.18]$; the largest effects in the model), and receptivity declines modestly with platform tenure, though the tenure intervals are marginal (Table~\ref{tab:regression}). The two most diagnostic effects concern need: users whose matches already convert to conversations are \emph{less} receptive ($\beta \approx -0.17$ on both dimensions), while users who report negative affect about stalled matches are \emph{more} receptive ($\beta \approx +0.14$). Receptivity concentrates where the friction is: it is highest among users reporting that the human-authored funnel is failing them. These are cross-sectional associations: the design does not license a causal reading, but the pattern is exactly what a need-based account predicts. This has a double edge. It is consistent with agents addressing a real, self-identified pain point rather than a manufactured one; it equally implies that the users most likely to adopt are those in the most emotionally loaded position, a targeting question we return to in the ethics discussion. The language coefficient is near zero on both dimensions, consistent with the pooling of the two samples.

\subsection{Measurement invariance}
Per-item likelihood-ratio DIF tests (Bonferroni-corrected) reject full cross-language invariance for two items (Y4: LRT~$=15.4$, $p=.001$; Y5: LRT~$=41.3$, $p<10^{-7}$); the remaining five are invariant, and a partial-invariance model freeing Y4/Y5 for the EN group leaves all headline quantities essentially unchanged (gap $= \gappartial$, $\rho = 0.92$, gender effects within $0.005$ of the constrained fit). By gender, only the agent-to-agent item shows DIF (Y5: LRT~$=16.6$); the two items defining the asymmetry (Y3, Y4) function equivalently for men and women, so the gender gap reflects trait differences rather than differential item functioning.

\emph{Where the invariance breaks, and what it may mean.} The non-invariance is not randomly located: both flagged items are \emph{receive}-side, while every send-side item, including deployment of one's own agent, functions equivalently across the two language forms. The freed parameters show the direction. For both items the lowest threshold sits substantially higher in the EN form (Y4: $b_1 = 0.47$ vs.\ $0.31$; Y5: $b_1 = -0.94$ vs.\ $-1.73$): at equal latent receptivity, EN-form respondents more often select the categorical-rejection options (``weird -- only want a real person''; ``strange and frightening''). Simultaneously, the EN group's latent receive-side mean rises to $+0.20$ under partial invariance: two offsetting effects that the fully constrained model conflated into a near-zero language coefficient (CI $[-0.09, 0.22]$), and that a sum-score analysis would never have separated. One reading is substantive: delegating is a largely private utility judgment (time saved, control retained), whereas accepting agent-mediated contact is a \emph{normative} one, where a judgment about politeness, authenticity, and what counts as strange, and norms are exactly what varies across linguistic--cultural contexts, consistent with the socially constructed character of authenticity judgments in AI-mediated communication \cite{hancock2020ai,ante2026cyrano}. We state this as a hypothesis rather than a finding: with 232 EN respondents self-selecting the English form on a single platform, language is confounded with unmeasured population composition, and translation artifacts in the rejection options' emotional valence cannot be separated from genuine norm differences. What the analysis does establish is methodological: receive-side receptivity is where measurement travels least well, so cross-locale deployments of the audit must invariance-test per market, and calibrate receive-side routing thresholds per locale rather than assume a single calibration. Cross-cultural replication with balanced samples is the natural extension, and the receive side is where variation should be sought.

\subsection{User segments}
\label{sec:lca}
A latent class analysis over the seven items: BIC-selected at four classes (27{,}632, vs.\ 28{,}031 at $K{=}3$ and 27{,}633 at $K{=}5$), with relative entropy $0.83$ and all 20 random restarts recovering the identical solution (mean ARI $= 1.00$), gives the population structure behind the continuous traits. The four- and five-class solutions are nearly tied on BIC ($\Delta \approx 1$); the choice is immaterial for our claims, because under $K{=}5$ the asymmetric-delegator class persists essentially unchanged (98.9\% of its members map to a $K{=}5$ class with the same profile and share, the extra class instead splitting the ambivalent group), as do the rejector and enthusiast classes. \emph{Rejectors} ($\approx$31\%) are negative on every item; women are overrepresented here (36\% of women vs.\ 29\% of men). \emph{Enthusiasts} ($\approx$19\%) are positive throughout, including toward receiving agents. \emph{Ambivalents} ($\approx$25\%) sit near the population mean with lukewarm responses across the board. The remaining class ($\approx$26\%) is the structurally consequential one: \emph{asymmetric delegators}, who overwhelmingly want their own agent (98\% respond ``want to try'' or ``might try''), enjoy the spectacle of agent-to-agent conversation, yet only 10\% would fully engage an agent that contacts them, and they rate overall product value low. This is the send/receive gap embodied as a population segment, roughly a quarter of the market wants to emit agent traffic it would not itself accept, and it is the segment a reciprocity policy prices out (\S\ref{sec:market}).

\subsection{Passive features versus active delegation}
The companion instrument (N = \Ngen) shows how specific this resistance is to \emph{delegation}. Among respondents using the closed response options (a minority left free-text comments instead, tabulated separately), passive generative features draw broad interest: an AI-written couple description reaches 80.1\% top-two-box interest and an AI profile summary 78.6\% (and 70.6\% say such features would raise their browsing interest). The one assistive feature that touches live conversation - AI communication tips - drops to 62.4\%, with 37.6\% explicitly uninterested; it is also the only feature men rate higher than women (65.1\% vs.\ 58.3\% top-two-box, against a female advantage on both passive features), consistent with the unmet-need pattern of \S\ref{sec:results}. Reception of generative AI in this population is therefore not monolithic aversion; it is graded by how deeply the feature intrudes into interpersonal communication, with autonomous delegation at the far end of the gradient.

\subsection{The routing signal predicts out of sample}
\label{sec:oos}
The routing counterfactual of \S\ref{sec:market} ranks receivers by receive receptivity and asks how engagement changes when agent traffic is routed to the most receptive. If receptivity were scored using the engagement item itself, that exercise would be partly circular: Y4 would help construct the score that then ``predicts'' Y4. We therefore validate the signal with the target item held out entirely. Under five-fold respondent-level cross-validation, we refit the two-dimensional model on the training folds \emph{excluding Y4}, score each held-out respondent's receive receptivity by EAP from the six non-target items and covariates only, calibrate a logistic link from score to Y4 on the training folds, and evaluate on the held-out folds. Pooled out-of-fold performance: AUC $= 0.89$ for any engagement (Y4 $\ge 1$) and $0.88$ for full engagement (Y4 $= 2$); Brier scores $0.132$ and $0.079$ against reference (base-rate) values of $0.240$ and $0.107$. Ranked into out-of-fold quartiles, actual full-engagement rates run $0.0\%$ / $2.4\%$ / $8.2\%$ / $37.9\%$ from least to most receptive quartile (any-engagement: $0.3\%$ / $18.4\%$ / $59.1\%$ / $81.4\%$). The top-quartile lift over the population base rate is $3.1\times$ (strict) - essentially the in-sample routing gain, now earned without the target item. A short instrument, minus the very item that defines engagement, ranks receivers well enough to triple per-contact engagement; this supports treating receive receptivity as a \emph{pre-deployment} routing signal (validation against behavior under live agent contact remains future work).

\begin{table}[t]
\caption{Routing-score baselines on identical folds (full engagement, out of fold). The logistic baselines are trained on the target labels; the latent score never uses Y4.}
\label{tab:baselines}
\small
\begin{tabular}{lcc}
\toprule
Score & AUC & Top-quartile lift \\
\midrule
Covariates only (logistic) & 0.60 & 1.5$\times$ \\
Raw $\mathrm{Y5{+}Y6}$ sum score & 0.85 & 3.0$\times$ \\
Logistic on six non-target items & 0.89 & 3.2$\times$ \\
Latent (EAP, leave-Y4-out) & 0.88 & 3.1$\times$ \\
\bottomrule
\end{tabular}
\end{table}
\emph{Against simpler scores.} Table~\ref{tab:baselines} benchmarks the latent score on the identical folds. Covariates alone barely beat chance (AUC 0.59): receive receptivity is not reducible to demographics or platform experience. A raw Y5{+}Y6 sum score reaches 0.85; a plain logistic on the six non-target items reaches 0.89. The latent score sits at 0.88, which is predictive parity with the supervised classifier. One asymmetry is worth noting: the logistic baselines are trained on the target labels in every training fold, while the latent score is estimated without ever observing Y4 (the logistic link fitted afterwards is monotone, so it does not affect ranking metrics). We claim parity, not superiority, over supervised classifiers, while additionally delivering what prediction-only scores cannot: threshold locations (the asymmetry itself), formal invariance testing, calibrated uncertainty, and a construct that transfers across instruments and markets.

\section{Stated-Preference Market Counterfactuals}
\label{sec:market}
The measurement model yields, for every respondent, model-implied \emph{endorsement} propensities: $p^{\mathrm{dep}}_i$, the probability of endorsing deployment of one's own agent (from item Y3 at $\hat\theta^{\mathrm{send}}_i$), and $p^{\mathrm{eng}}_i$, the probability of endorsing engagement with a counterpart's agent (from Y4 at $\hat\theta^{\mathrm{recv}}_i$). These are stated-preference quantities, not calibrated behavior; we report a \emph{strict} operationalization (top response category only) and a \emph{soft} one (half-credit for the intermediate ``might try'' / ``might engage'' category) and read the pair as scenario bounds. Population means already reveal the imbalance: deployment 0.38/0.50 (strict/soft) against engagement 0.12/0.26.

We then compute a random-pairing counterfactual over directed dyads: sender $i$ initiates an agent contact with probability $p^{\mathrm{dep}}_i$, receiver $j$ engages with probability $p^{\mathrm{eng}}_j$, representing an independence baseline, not an equilibrium matching model. Two outcomes matter to a platform. Under a sender-eligibility rule $S$ and a receiver-routing rule $R$ (subsets of the population; the unconstrained market has $S$ and $R$ equal to everyone), the \emph{viable dyad rate} measures volume,
\begin{equation}
V(S,R) \;=\; \mathbb{E}_{i}\!\left[p^{\mathrm{dep}}_i \mathbf{1}\{i \in S\}\right]\,
\mathbb{E}_{j}\!\left[p^{\mathrm{eng}}_j \mathbf{1}\{j \in R\}\right],
\end{equation}
and \emph{engagement per agent contact} measures quality, $Q(R) = \mathbb{E}_{j \in R}[\,p^{\mathrm{eng}}_j\,]$. Every design lever in this section is a choice of $(S,R)$, so its price is a closed-form computation on fitted quantities in stage four of the audit. Table~\ref{tab:market} reports both outcomes under four regimes.

\begin{table}[t]
\caption{Random-pairing stated-preference counterfactuals. Vol.\ = viable dyad rate; Qual.\ = engagement per agent contact.}
\label{tab:market}
\small
\begin{tabular}{lrrrr}
\toprule
 & \multicolumn{2}{c}{Strict} & \multicolumn{2}{c}{Soft} \\
\cmidrule(lr){2-3}\cmidrule(lr){4-5}
Scenario & Vol. & Qual. & Vol. & Qual. \\
\midrule
Baseline (unconstrained) & 0.044 & 0.116 & 0.128 & 0.256 \\
Male $\to$ female contacts & 0.035 & 0.084 & 0.114 & 0.213 \\
Female $\to$ male contacts & 0.042 & 0.134 & 0.122 & 0.279 \\
Reciprocity gate & 0.020 & 0.116 & 0.046 & 0.256 \\
Routing: top 10\% receivers & 0.025 & 0.648 & 0.041 & 0.809 \\
Routing: top 25\% receivers & 0.037 & 0.394 & 0.080 & 0.640 \\
Routing: top 50\% receivers & 0.043 & 0.227 & 0.117 & 0.469 \\
Routing: top 75\% receivers & 0.044 & 0.154 & 0.127 & 0.338 \\
\bottomrule
\end{tabular}

\end{table}

\textbf{Baseline.} Unconstrained, only 4.4\% (strict) to 12.8\% (soft) of directed dyads yield an engaged agent interaction, and engagement per contact is 11.6\%/25.6\%: most modeled agent contacts do not result in endorsed engagement. Stated enthusiasm for \emph{having} an agent coexists with a market in which most of what agents would produce is, by the recipients' own statements, unwanted.

\textbf{Direction matters.} Because women are less receptive on both dimensions, the market is directionally imbalanced: male-deployed agents contacting women achieve 8.4\% engagement per contact (strict) against 13.4\% for female-deployed agents contacting men. The heavier prospective traffic (male-to-female) faces the colder reception: the same structural pattern that plagues human-authored first messages, reproduced and potentially amplified by automation.

\textbf{Reciprocity gate.} Requiring symmetric willingness, when a user may deploy an agent only if their own soft engagement propensity is at least $0.5$, excludes 65\% of likely deployers and halves interaction volume (0.044~$\to$~0.020 strict). This prices the free-rider structure directly: about a quarter of users want to send agents but will not receive them, and a reciprocity norm removes exactly this traffic.

\begin{figure}[t]
\includegraphics[width=\linewidth]{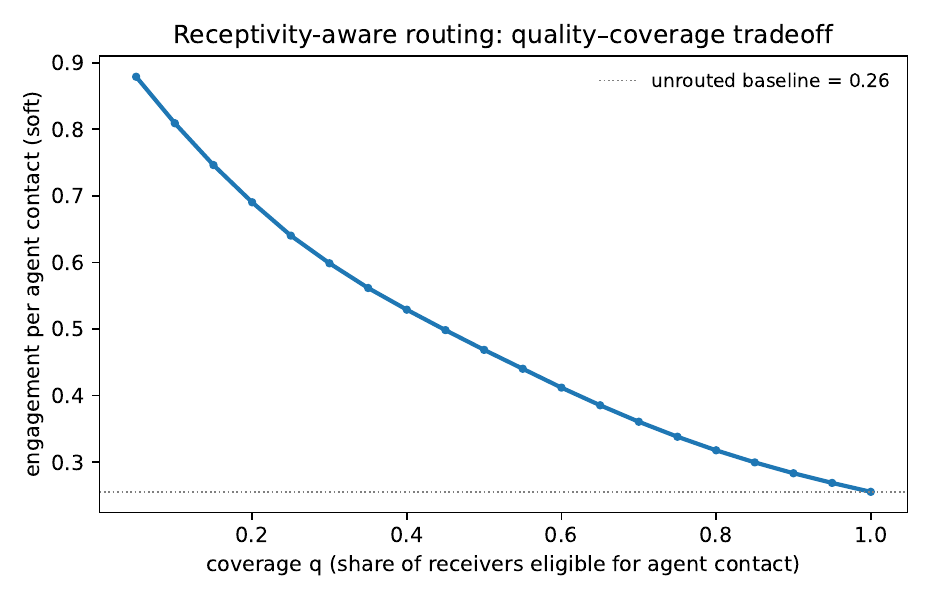}
\caption{Receptivity-aware routing: engagement per agent contact (soft operationalization) as a function of receiver coverage $q$. The dotted line marks the unrouted baseline.}
\label{fig:routing}
\end{figure}
\textbf{Receptivity-aware routing.} Routing agent contacts only to receivers in the top receptivity quartile raises engagement per contact from 11.6\% to 39.4\% (strict; 25.6\%~$\to$~64.0\% soft), or a 3.4$\times$ quality gain, at the cost of restricting coverage to 25\% of receivers. This is not an artifact of scoring receivers with the engagement item itself: with Y4 held out and receptivity re-estimated under cross-validation, the top-quartile lift is $3.1\times$ (\S\ref{sec:oos}). Figure~\ref{fig:routing} traces the full quality--coverage frontier, which a platform can tune like any exposure policy. Routing on \emph{receive} receptivity is a matching-layer intervention invisible to senders; it converts the measurement model into a pre-deployment policy input.

Two caveats bound these numbers. The simulation assumes random dyad formation, ignoring matching structure and homophily in receptivity. It treats propensities as static, whereas receivers plausibly update after good or bad agent encounters, reflecting that an equilibrium version is future work. The qualitative conclusions: receive-side scarcity, directional imbalance, and the volume/quality lever structure, do not depend on these simplifications.

\section{Design Implications for Agentic Recommenders}
\textbf{Receive-side consent is a first-class primitive.} Current agent designs treat deployment as the consent event: the sender opts in. Our results locate the scarce resource on the other side. A receiving-consent surface, or an explicit setting for whether, and from whom, agent-mediated contact is acceptable, is not a compliance nicety but the mechanism that determines market quality; without it, three quarters of agent contacts land on users who did not want them (Table~\ref{tab:market}). Disclosure follows directly: undisclosed agents convert unwilling receivers into deceived ones, the ``digital betrayal'' documented qualitatively by \cite{ante2026cyrano}, and our receive-side estimates should be read as an \emph{upper} bound on tolerance for undisclosed contact.

\textbf{Receptivity is a routable signal.} Because receive receptivity is measurable from a handful of items and predicts held-out engagement responses with AUC $0.88$ even when the engagement item itself is excluded from scoring (\S\ref{sec:oos}), it can enter the matching layer like any exposure feature (in production, behavioral proxies would play the items' role). The quality--coverage frontier (Figure~\ref{fig:routing}) is the tuning knob: routing agent traffic to the top receptivity quartile more than triples per-contact engagement while leaving human-authored contact untouched for everyone else. This is the cheapest intervention in our counterfactual set and requires no visible product change for senders.

\textbf{Reciprocity is a values choice with a known price.} A symmetric-willingness rule (deploy only if you would engage) halves agent-interaction volume and excludes two thirds of would-be deployers - almost exactly the asymmetric-delegator segment. Whether to charge that price is a platform-values decision, not a technical one; our contribution is that the price is now quantified.

\textbf{Beyond dating: auditing delegated-communication markets.}\label{sec:beyond}
The audit transfers wherever delegation meets a human receiver. Recruiting is the nearest analogue: agent-authored candidate outreach is already deployed, candidate-side receptivity is unmeasured, and the market shares dating's structure of concentrated attention and scarce replies. Sales prospecting, agent-mediated customer contact, and professional networking follow the same pattern. In each, the audit's outputs map onto the same levers: deployment eligibility, receive-side routing (invariance-tested and calibrated per locale, per \S\ref{sec:results}), and disclosure mechanics. The same failure mode looms: send-side enthusiasm masking receive-side scarcity. We conjecture the delegation asymmetry itself generalizes, because its plausible mechanisms (retained control over one's own agent, none over others'; effort saved when sending, authenticity lost when receiving) are not dating-specific; testing that conjecture requires only re-fielding the instrument, which is the point of packaging the procedure.

\textbf{Sequence the rollout by intrusion depth.} The passive-to-active gradient (\S\ref{sec:results}) suggests a staged path: profile-level generative features first (75\%+ interest, low relational intrusion), conversation assistance with strong user control second, autonomous delegation last and opt-in on both sides. Deploying the far end of the gradient first, to the most frustrated users, is the demand-efficient and trust-corrosive order.

\section{Limitations}\label{sec:limits}
Our estimates are stated preferences elicited from concept mockups, not behavior under deployed agents; the technology-acceptance literature suggests stated and revealed adoption correlate but diverge, and the direction of divergence for \emph{receiving} agents is unknown. Both samples are self-selected respondents from a single platform, and the EN subsample is small (n = 232) and self-selected, so language is confounded with population composition. Our cross-lingual claims are therefore limited to the partial-invariance form we report, and the cultural interpretation of the receive-side DIF remains a hypothesis. Ordinal codings required judgment calls for response options without a canonical order; the sensitivity analyses show the headline quantities are robust to the defensible alternatives, but cannot rule out orderings we did not consider. The two instruments were fielded to non-linked respondents, so the passive-versus-active contrast is between-sample. The market simulation assumes random dyad formation and static propensities, abstracting from matching structure, homophily in receptivity, and the plausible dynamics in which receivers update after agent encounters. Finally, attitudes toward agent mediation are likely nonstationary as the technology normalizes; our estimates are a 2026 snapshot of a moving object.

\section{Conclusion}
Agent-mediated matching is being built sender-first, but it will succeed or fail receiver-first. Measuring both sides of delegation within person, we find distinct constructs separated by a robust 0.7~SD asymmetry, a quarter of users wanting to send agent traffic they would not fully accept, and, under a random-pairing counterfactual, only 4--13\% of directed dyads combining deployment with engagement unless receive-side consent becomes a design primitive. The measurement model that produces these numbers is cheap with only seven ordinal items, and its pre-deployment outputs are directly usable, from a validated routing signal to the quantified price of a reciprocity norm. As LLM agents enter interpersonal products, we offer this as a template for measuring readiness and a caution against mistaking enthusiasm for delegation as enthusiasm for its receipt.

\section*{Ethical Considerations}
\textbf{Data handling and review.} Both surveys were voluntary, fielded by the platform to its users for product research, and provided to the authors as existing data; this study is a secondary analysis of de-identified data, determined by the authors' institutional review board not to constitute human subjects research (protocol 26-08-10287). Analyses are reported in aggregate: no individual-level data, no free-text excerpts attributable to individuals, and no demographic cells small enough to risk re-identification (a concern for the small EN subsample in particular). Contact information volunteered by interview candidates was excluded from analysis. The public deposit (\S\ref{sec:data}) contains only the anonymized structured responses: direct identifiers and free-text are removed, timestamps are coarsened, and demographic cells were audited so that no released cross-classification isolates fewer than $k=5$ respondents; raw responses are not released.

\textbf{Dual use of receptivity scoring.} A model that scores users' receptivity to AI contact can protect users (routing agent traffic away from those who do not want it) or exploit them (segmenting users for maximal AI exposure regardless of preference). The unmet-need association sharpens this: the most receptive users are those most frustrated with their current experience, so demand-efficient targeting concentrates an experimental technology on users in an emotionally vulnerable position. We consider receive-side routing defensible precisely because it acts on the \emph{receiving} user's own preference; deployment-side targeting of frustrated users optimizes the platform's adoption curve against the user's state, and we caution against it.

\textbf{Gender differences.} We report robust gender differences in receptivity, verified to reflect trait differences rather than differential item functioning. These describe aggregate distributions and must not be used to gate features by gender or to justify differential treatment of individuals; their design-relevant content is directional (the heaviest prospective agent traffic faces the least receptive audience), not individual.

\textbf{Deception and disclosure.} Undisclosed agent communication converts our receive-side estimates from consent measurements into deception measurements: users who would decline agent contact cannot decline what they cannot detect. Our results therefore argue for mandatory disclosure of agent authorship, and we note that prior work finds trust penalties even for \emph{suspected} AI authorship, so disclosure is also in platforms' long-run interest.

\textbf{Agent-mediated intimacy.} Delegating early romantic communication to agents raises questions beyond any platform: whose words form the basis of a relationship, and what happens at the transition to unassisted interaction. Our data show users themselves articulate these concerns unprompted. We present measurement and market analysis to inform this debate, not to settle it.

\bibliographystyle{ACM-Reference-Format}
\bibliography{references}

\end{document}